\documentclass[11pt,a4paper]{article}
\usepackage[hyperref]{style/naacl2021}
\usepackage{times}
\usepackage{inconsolata}
\usepackage{amssymb}
\usepackage{pifont}

\usepackage{url}
\usepackage{bibunits}
\newif\ifcomment
\commenttrue
\commentfalse

\newif\ifdoublespaceme
\doublespacemefalse

\usepackage{framed}
\usepackage{soul}
\usepackage{mdwlist}
\usepackage{latexsym}
\usepackage{nicefrac}
\usepackage{booktabs}
\usepackage{amsfonts}
\usepackage{bold-extra}
\usepackage{amsmath}
\usepackage{fnpct}
\usepackage{dsfont}
\usepackage{amssymb}
\usepackage{bm}
\usepackage{graphicx}
\usepackage{mathtools}
\usepackage{microtype}
\usepackage{multirow}
\usepackage{multicol}
\usepackage{xspace} 
\usepackage{comment}
\usepackage{subfigure}

\usepackage{xcolor}
\usepackage[nomessages]{fp}
\usepackage{tikz}
\usepackage{calc}
\usepackage{pgfplots, pgfplotstable}
\usepackage{filecontents}

\pgfplotstableset{col sep=comma}

\newcommand{\gem}[1]{\mbox{\textsc{gem}}}
\newcommand{\abr}[1]{\textsc{#1}}

\newcommand{\hidetext}[1]{}
\newcommand{\ignore}[1]{}

\ifcomment
\newcommand{\todo}[1]{\textcolor{red}{{\bf TODO: #1}}}
\else
\newcommand{\todo}[1]{}
\fi

\ifcomment
\newcommand{\pinaforecomment}[3]{\colorbox{#1}{\parbox{.8\linewidth}{#2: #3}}}
\else
\newcommand{\pinaforecomment}[3]{}
\fi

\newcommand{\jbgcomment}[1]{\pinaforecomment{red}{JBG}{#1}}

\newcommand{\smallurl}[1]{ \begin{tiny}\url{#1}\end{tiny}}

\definecolor{lightblue}{HTML}{3cc7ea}
\definecolor{CUgold}{HTML}{CFB87C}
\definecolor{grey}{rgb}{0.95,0.95,0.95}
\definecolor{ceil}{rgb}{0.57, 0.63, 0.81}

\newlength\maxlen
\newlength\unitlen

\definecolor{exampleblue}{RGB}{189, 215, 238}
\definecolor{exampletextblue}{RGB}{47, 85, 151}

\makeatletter
\renewcommand\sectionautorefname{\S\@gobble}  

\newcommand{\figfile}[1]{2021_naacl_multi_ance/figures/#1}

\newcommand{\hotpot}{\textsc{HotPotQA}\xspace} 
\newcommand{\bert}{{\abr{bert}}\xspace} 
\newcommand{\name}{\textsc{BeamDR}\xspace} 
\newcommand{\grr}{\textsc{\abr{grr}}\xspace} 
\newcommand{\fone}{$F_1$}

\title{Multi-Step Reasoning Over Unstructured Text with Beam Dense Retrieval}

\author{
  Chen Zhao\\
  University of Maryland\\
  {\tt chenz@cs.umd.edu} \\\And
  Chenyan Xiong\\
  Microsoft Research\\
  {\tt chenyan.xiong@microsoft.com} \\\AND
  Jordan Boyd-Graber\\
  University of Maryland\\
  {\tt jbg@umiacs.umd.edu} \\\And
  Hal Daum{\'e} III\\
  Microsoft Research \& University of Maryland\\
  {\tt me@hal3.name} \\}

\date{}

\begin{document}
\maketitle

\ifdoublespaceme
  \doublespacing
\fi

\begin{abstract}

Complex question answering often requires finding
a reasoning chain that consists of 
multiple evidence pieces.
%
Current approaches incorporate the strengths 
of structured knowledge and unstructured text,
assuming text corpora is semi-structured. 
%
Building on dense retrieval methods,
we propose a new multi-step retrieval approach (\name{})
that iteratively forms an evidence chain through beam search in 
dense representations.
When evaluated on multi-hop question answering, \name{} is competitive
to state-of-the-art systems, \emph{without} using any semi-structured
information.
Through query composition in dense space, 
\name{} captures the implicit relationships between evidence 
in the reasoning chain.  The code is available at \url{https://github.com/henryzhao5852/BeamDR}.

\end{abstract}


\section{Introduction}
\label{sec:intro}

\begin{figure}[!t]
    \centering
    \includegraphics[width=0.9\linewidth]{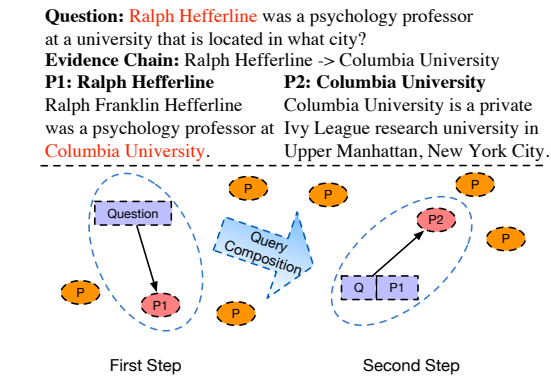}
    \caption{Top: A complex question example from \hotpot{} that
      requires finding an evidence chain. Bottom: \name{} iteratively
      composes the new query and retrieves evidence in dense space
      without the need for linked documents.  }
    \label{fig:ex}
    \end{figure}

Answering complex questions requires combining knowledge pieces
through multiple steps into an evidence chain (\underline{Ralph
  Hefferline} $\rightarrow$ \underline{Columbia University} in
Figure~\ref{fig:ex}).
When the available knowledge sources are graphs or databases, 
constructing chains can use the sources' inherent structure.
However, when the information needs to be pulled from unstructured
text (which often has better coverage), standard information retrieval
(\abr{ir}) approaches only go ``one hop'': from a query to a
single passage.

Recent approaches~\cite[\emph{inter alia}]{Dhingra2020Differentiable, Zhao:Xiong:Qian:Boyd-Graber-2020, zhaotransxh2020,asai2020learning} try to 
achieve the best of both worlds: use the
unstructured text of Wikipedia with its structured hyperlinks.
While they show promise on benchmarks, it's difficult to
extend them beyond academic testbeds because real-world
datasets often lack this structure.
For example, medical records lack links between reports.

Dense retrieval~\cite[\emph{inter alia}]{lee-etal-2019-latent,
  guu2020realm, karpukhin-etal-2020-dense} provides a promising path to
overcome this limitation.
It encodes the query and evidence (passage) into dense vectors and matches them in the embedding space.
In addition to its efficiency---thanks to maximum inner-product search
(\abr{mips})---\newcite{xiong2020approximate} show that dense
retrieval rivals
\bert{}~\cite{devlin+19}-based (sparse) retrieve-then-rerank \abr{ir}
pipelines on single step retrieval.
Unlike traditional term-based retrieval, fully learnable dense encodings 
provide flexibility for different tasks.

This paper investigates a natural question: can we build a 
retrieval system to find an evidence chain on unstructured text corpora?
We propose a new multi-step dense retrieval method to model the
implicit relationships between evidence pieces.
We use beam search (Section~\ref{sec:model}) in the dense
space to find and cache the most relevant candidate chains and 
iteratively compose the query by appending the retrieval history.
We improve the retrieval by encouraging the representation to
discriminate hard negative evidence chains from the correct chains,
which are refreshed by the model.

We evaluate \textbf{Beam} \textbf{D}ense \textbf{R}etrieval (\name{}) 
on \hotpot{}~\cite{yang+18b}, a multi-hop question answering benchmark. 
When retrieving evidence chains directly from the corpus (full retrieval), \name{} 
is competitive to the 
state-of-the-art cascade reranking systems that use Wikipedia links. Combined with standard 
reranking and answer span extraction modules,
the gain from full retrieval propagates to findings answers
(Section~\ref{sec:experiment}). 
By iteratively composing the query representation, \name{} captures
the hidden ``semantic'' relationships in the evidence
(Section~\ref{sec:analysis}).



\section{\name{}: Beam Dense Retriever}
\label{sec:model}

This section first discusses preliminaries 
for dense retrieval, then 
introduces our method, \name{}.

\subsection{Preliminaries}

Unlike classic retrieval techniques, dense retrieval methods match
distributed text representations~\cite{bengio-13} rather than sparse
vectors~\cite{salton-68}.
With encoders (e.g., \bert{}) to embed query~$q$ and passage~$p$ into
dense vectors~$E_{Q}(q)$ and~$E_{P}(p)$, the relevance score $f$ is
computed by a similarity function $sim(\cdot)$ (e.g., dot product)
over two vector representations:
\begin{equation}
    f(q, p) = sim(E_{Q}(q), E_{P}(p)).
    \label{eqn:sim}
  \end{equation}
  After encoding passage vectors offline, we can efficiently retrieve
  passage through approximate nearest neighbor search over the maximum
  inner product with the query, i.e.,
  \abr{mips}~\cite{shrivastava2014asymmetric,JDH17}.

\subsection{Finding Evidence Chains with \name{}}
\label{subsec:beamdr}

We focus on finding an evidence chain from an unstructured text corpus
for a given question, often the hardest part of complex question
answering.
We formulate it as multi-step retrieval problem.
Formally, given a question~$q$ and a corpus $C$, the task is
to form an ordered evidence chain~${p_{1}...p_{n}}$ from~$C$, with
each evidence a passage.
We focus on the supervised setting, where the labeled evidence set
is given during training (but not during testing).

Finding an evidence chain from the corpus 
is challenging because:
1) passages that do not share enough words are hard to retrieve 
(e.g., in Figure~\ref{fig:ex}, the evidence \underline{Columbia University});
2) if you miss one evidence, you may err on all
that come after.

We first introduce scoring a single evidence chain, then finding the
top $k$ chains with beam search, and finally training \name{}.
%
%


\paragraph{Evidence Chain Scoring}

The score $S_{n}$ of evidence chain ${p_{1},\dots,p_{n}}$ is the product of the
(normalized) relevance scores of individual evidence pieces.
At each retrieval step $t$, 
to incorporate the information from both the question and retrieval history, we compose  
a new query~$q_{t}$ by appending the tokens of retrieved chains $p_{1},\dots,p_{t - 1}$ 
to query $q$ ($q_t=[q;p_{1};\dots;p_{t-1}]$), we use \abr{mips} 
to find relevant evidence piece $p_t$ from the corpus and update the
evidence chain score $S_t$ by multiplying the
current step $t$'s relevance score $f (q_{t}, p_{t}) * S_{t -1}$.

\paragraph{Beam Search in Dense Space}

Since enumerating all evidence chains is computationally impossible,
%
we instead maintain an evidence cache.
In the structured search literature this is called a \emph{beam}:
the $k$-best scoring candidate chains we have found thus far.
%
%
%
We select evidence chains with beam search in dense space.  At step
$t$, we enumerate each candidate chain $j$ in the beam $p_{j,
  1}...p_{j, t - 1}$, score the top $k$ chains and update the beam.
After $n$ steps, the $k$ highest-scored evidence chains with
length~$n$ are finally retrieved.

\paragraph{Training \name{}}

The goal of training is to learn embedding functions that
differentiate positive (relevant) and negative evidence chains.
Since the evidence pieces are unordered, we use heuristics to infer the order of evidence chains.
A negative chain has at least one evidence piece that is not in the gold 
evidence set. 
For each step $t$, the input is
the query~$q$, a positive chain~$P_{t}^{+} = p^{+}_{1},\dots,p^{+}_{t}$
and~$m$ sampled negative chains $P_{j, t}^{-} = p^{-}_{1},\dots,p^{-}_{t}$. 
We update the negative log likelihood (\abr{nll}) loss:
\begin{align}
  &L(q, P^{+}, P_{1}^{-}, ..., P_{m}^{-})   \\
  &=\sum_{t}{\frac{e^{f([q;P_{t-1}^{+}], p^{+}_{t})}}{e^{f([q;P_{t-1}^{+}], p^{+}_{t})} + \sum_{j=1}^{m} {e^{f([q;P_{j, t-1}], p^{-}_{j, t})}}}}.\nonumber
  \label{eqn:loss}
\end{align}
Rather than using local in-batch or term matching negative samples,
like \citet{guu2020realm} we select negatives from the whole corpus,
which can be more effective for single-step
retrieval~\cite{xiong2020approximate}.
In multi-step retrieval, we select negative evidence chains from the corpus.
Beam search on the training data finds the top~$k$
highest scored negative chains for each retrieval step. Since the model parameters are dynamically updated, 
we asynchronously refresh the negative chains with 
the up-to-date model checkpoint~\cite{guu2020realm, xiong2020approximate}. 

\section{Experiments: Retrieval and Answering}
\label{sec:experiment}

\jbgcomment{Can we have more descriptive section titles?}

Our experiments are on \hotpot{} fullwiki setting~\cite{yang+18b}, the
multi-hop question answering benchmark.  We mainly evaluate on
retrieval that extracts evidence chains (passages) from the corpus; we
further add a downstream evaluation on whether it finds the right
answer.

\subsection{Experimental Setup}

\paragraph{Metrics}
Following \citet{asai2020learning}, we report four metrics
on retrieval:
answer recall (\abr{ar}), if answer span is in the retrieved
passages;
passage recall (\abr{pr}), if at least one gold passage is in the
retrieved passages;
Passage Exact Match (\abr{p em}), if both gold passages are included
in the retrieved passages;
and Exact Match (\abr{em}),
whether both gold passages are included in the top two retrieved
passages (top one chain).
We report exact match (\abr{em}) and \fone{} on
answer spans.

\paragraph{Implementation}

We use a \bert{}-base encoder for retrieval and report both \bert{}
base and large for span extraction.
We warm up \name{} with \abr{tf-idf} negative chains.
The retrieval is evaluated on ten passage chains (each chain has two
passages).
To compare with existing retrieve-then-rerank cascade systems, we
train a standard \bert{} passage reranker~\cite{nogueira2019passage},
and evaluate on ten chains reranked from the top 100 retrieval
outputs.
We train \name{} on six 2080Ti GPUs, 
three for training, three for refreshing negative chains. 
We do not search hyper-parameters and use suggested ones 
from \citet{xiong2020approximate}.
%

\subsection{Passage Chain Retrieval Evaluation}
\begin{table}[!tb]
\small
  \vspace{-0.3cm}
  \footnotesize
  \begin{tabular}{m{3cm}	m{0.6cm} m{0.6cm} m{0.85cm} m{0.6cm} }\toprule 
   Models  & AR & PR & P EM & EM \\
    \midrule
    \multicolumn{5}{l}{\emph{Full Retrieval} }\\
    \abr{tf-idf} & 39.7 & 66.9  &10.0 & 18.2 \\
    
    \abr{mdr} $^*$ & 75.4 & - & 65.9 & -\\ 
    \name{} (\abr{ir} Neg) & 76.8 & 86.4 & 64.1 & 40.4\\
    \name{} (Greedy) & 83.6 & 90.7 & 72.7 & 34.1\\
    \name{} (Ours) & 87.0 & 92.9 & 79.2 & 60.7\\ \hline
    \multicolumn{5}{l}{\emph{Reranking from Retrieval Outputs}}\\
    \abr{sr} & 77.9 & 93.2 & 63.9 & 46.5\\ 
      \grr{}   & 87.8  & 93.3 & 77.9 & 61.1 \\
      \abr{mdr} $^*$ & 88.2 & - & 81.2 & -\\ 
      \name{} (Ours) & 90.7 & 94.7 & 83.7 & 70.7\\ 
    \bottomrule
  \end{tabular}
      \caption{Compare \name{} with other retrieval systems.
      Top: Retrieval from the whole corpus, bottom: Reranking from top 100 full retrieval outputs. $^*$ indicates parallel work.}
      
      \label{tb:retrieval_results}
  
  \end{table}

\paragraph{Baselines}
We compare \name{} with \abr{tf-idf}, Semantic
Retrieval~\cite[\abr{sr}]{nie2019revealing}, which uses a cascade \bert{}
pipeline, and the Graph recurrent
retriever~\cite[\grr{}]{asai2020learning}, our main baseline, which
iteratively retrieves passages following the Wikipedia hyperlink
structure, and is state-of-the-art on the leaderboard.
We also compare against a contemporaneous model, multi-hop dense
retrieval~\cite[\abr{mdr}]{xiong2020answering}.

%
  
   


\paragraph{Results: Robust Evidence Retrieval without Document Links}

Table~\ref{tb:retrieval_results} presents retrieval results. 
On full retrieval, 
\name{} is competitive to \abr{grr}, state-of-the-art \emph{reranker} 
using Wikipedia hyperlinks.
\name{} also has better retrieval than the contemporaneous \abr{mdr}.
Although both approaches build on dense retrieval, \abr{mdr}
is close to \name{} with \abr{tf-idf} negatives. We instead refresh
negative chains with intermediate representations, which help the model
better discover evidence chains.
Our ablation study (Greedy search) indicates the importance of maintaining the beam during inference.  
With the help of cross-attention between the question and the passage, 
using \bert{} to rerank \name{} outperforms all
baselines.

\paragraph{Varying the Beam size}

\begin{figure}[t]
  \begin{center}
  \includegraphics[width=0.9\linewidth]{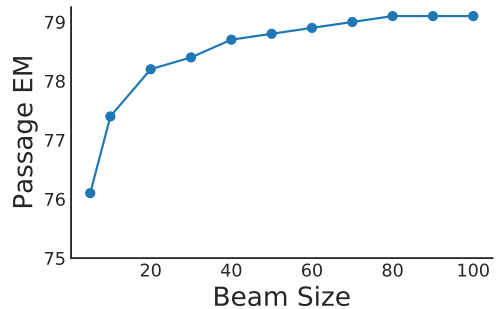}
  \end{center}
  \caption{Passage retrieval accuracy on different beam size. Our system is robust to the increase of beam size.}
  \label{fig:beamsize}
\end{figure}

Figure~\ref{fig:beamsize} plots the Passage \abr{em} with different
beam sizes.  While initially increassing the beam size improves
Passage Exact Match, the marginal improvement decreases after a beam
size of forty.

\subsection{Answer Extraction Evaluation}

\paragraph{Baselines} We compare \name{} with  
 \abr{txh}~\cite{zhaotransxh2020}, \abr{grr}~\cite{asai2020learning} and 
 the contemporaneous \abr{mdr}~\cite{xiong2020answering}. 
 %
  We use released code from \abr{grr}~\cite{asai2020learning} following its settings on
 \bert{} base and large. We use four 
 2080Ti GPUs.

\begin{table}
  \vspace{-0.3cm}
   \begin{tabular}{llcccc}\toprule 
      \multirow{2}{*}{Retriever}  &\multirow{2}{*}{Reader} &\multicolumn{2}{c}{Dev}& \multicolumn{2}{c}{Test}\\
      & & EM & F1 &  EM & F1  \\\hline
     \multicolumn{6}{l}{\emph{BERT base Reader} }\\
     \abr{txh} & \abr{txh}& 54.0 & 66.2 &51.6& 64.1  \\ 
     \grr{} & \grr{} & 52.7 & 65.8  & -& -   \\ 
     \name{}  & \grr{} & 54.9 & 68.0 &-&-  \\\hline
     \multicolumn{6}{l}{\emph{BERT large wwm Reader} }\\
     \grr{} & \grr{}& 60.5 & 73.3  &60.0&   73.0 \\ 
     \name{} & \grr{} & 61.3 &  74.1 &60.4&  73.2 \\ 
     \abr{mdr}$^*$ & \abr{mdr}$^*$& 61.5 & 74.7  &-&- \\ \hline
     \multicolumn{6}{l}{\emph{ELECTRA large Reader} }\\
     \abr{mdr}$^*$ & \abr{mdr}$^*$& 63.4 & 76.2 &62.3 & 75.3  \\


      \midrule
    \end{tabular}
        \caption{\hotpot{} dev and test set answer exact match (EM) and F1 results. $^*$ indicates parallel work.}
        
        \label{tb:dev_ans}
    
    \end{table}

\paragraph{Results}

Using the same implementation but on our reranked chains, \name{}
outperforms \grr{} (Table~\ref{tb:dev_ans}), suggesting gains from
retrieval could propagate to answer span extraction.
\name{} is competitive with \abr{mdr} but slightly lower; we speculate
different reader implementations might be the cause.

\section{Exploring How we Hop}
\label{sec:analysis}

In this section, we explore how \name{} constructs
evidence chains.

\begin{figure}[t]
    \centering
    \includegraphics[width=0.85\linewidth]{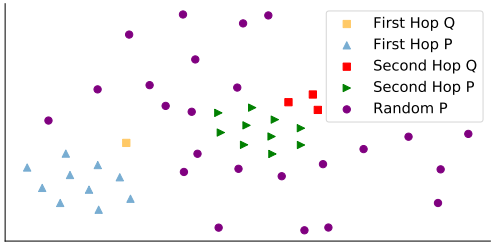}
    \caption{\abr{t-sne} visualization of query (Q) and passage (P)
      embeddings over different retrieval steps.  
      \name{} conducts multi-step reasoning by hopping in the learned representation space.}
    \label{fig:tsne}
\end{figure}


\begin{table}[t]
    \centering

    \footnotesize
        
    \begin{tabular}{lcc|c}
  
      \toprule
      \multirow{2}{*}{Models} & \multicolumn{2}{c|}{Passage Recall} & \multirow{2}{*}{Overlap}\\
        & First hop & Second hop & \\
       \midrule
       \grr{}   &85.1& 85.3  &64.3\\
       \name{}   & 86.4  & 78.9 & 26.7\\
       \name{}$	\dagger$   &  88.0 &  87.1 &14.7\\
      \bottomrule
    \end{tabular}
    \caption{Passage Recall and overlap comparison between \name{} and
      \grr{} with different hop passages. Systems with~$\dagger$
      filter second hop passages with links. }
        \label{tb:hops}

  \end{table}


\subsection{Qualitative Analysis}
Figure~\ref{fig:tsne} shows query and passage representations with
\abr{t-sne}~\cite{maaten2008visualizing}.
Unsurprisingly, in the dense space, the first hop query (question) is 
close to its retrieved 
passages but far from second hop passages (with 
some negative passages in between).
After composing the question and first hop passages, the second hop queries
indeed land closer to the second hop passages. 
Our 
quantitative analysis (Table~\ref{tb:hops}) further shows 
\name{} has little overlap between retrieved passages in two hops.
\name{} mimics multi-step reasoning by
hopping in the learned representation space.

\subsection{Hop Analysis}

To study model behaviors under different hops, we use heuristics\footnote{We label the passage that contains the answer
as the second hop passage, while the other one as the first hop passage. If both passages include the answer,
passage title mentioned in the question is the first hop passage.
} to infer the order of evidence passages.
In Table~\ref{tb:hops}, \name{} slightly wins on first hop passages, 
with the help of hyperlinks, 
\grr{} outperforms \name{} on second hop retrieval.
Only $21.9\%$ of the top-10 \name{} chains are
connected by links.  
\name{} wins
after using links to filter candidates.

\subsection{Human Evaluation on Model Errors and Case Study}

To understand the strengths and weaknesses of \name{} compared with
\grr{}, we manually analyze $100$ bridge questions from the \hotpot{}
development set. \name{} predicts fifty of them correctly and \grr{}
predicts the other fifty correctly (Tables~\ref{tb:human-eval} and~\ref{tb:case}).


\paragraph{Strengths of \name{}.}
Compared to \grr{}, the largest gain of \name{} is to identify question entity passages. 
As there is often little context overlap besides the entity surface form, 
a term-based approach (\abr{tf-idf} used by \grr{}) falters.
Some of the \grr{} errors 
also come from using reverse links to find second hop passages (i.e., the second hop passage links to the 
first hop passage). 

\begin{table}[!tb]
    \centering
    \small
      \begin{tabular}{lccc}
        \toprule
       \textbf{Errors} & \textbf{Type}  & \textbf{$\%$}\\
       \midrule
       & Question entities& 62  &\\
       \grr{} & Connect with reverse links& 16  &\\
       & Text matching& 14  &\\
       &Others& 8  &\\
       \midrule
       &Text matching& 46  &\\
       \name{} &No links between passages& 39  &\\
       & Question entities& 15  &\\
        
      \bottomrule
      \end{tabular}
      \caption{We manually analyze 100 bridge questions and categorize model errors. 
      }
    \label{tb:human-eval}

  \end{table}

\paragraph{Weaknesses of \name{}.}
Like \citet{karpukhin-etal-2020-dense}, many of \name{}'s
errors could be avoided by simple term matching. For example, matching 
\textit{``What screenwriter with credits for Evolution co-wrote a film \ul{starring Nicolas Cage and T\'ea Leoni}?''}
to the context \textit{``The Family Man is a 2000 American film written by David Diamond and David Weissman,
and \ul{starring Nicolas Cage and T\'ea Leoni}.''}.

\begin{table}[t]
  \centering

  \small
  \footnotesize

 \begin{tabular}{p{7.3cm}}
       \hline 
      \textbf{Q}: \underline{Chris Williams} last played for which football club from the National League North?

      \textbf{Passage 1}: \underline{Christopher Jonathan "Chris" Williams} 
      is an English semi-professional footballer who last played for \underline{Salford City} as a forward.
      
      \textbf{Passage 2}: \underline{Salford City Football Club} is a professional football club in the Kersal area of Salford, Greater Manchester, England.

       \textbf{\name{}}: Chris Williams (English Footballer) $\rightarrow$ Salford City F.C. \checkmark
      
      \textbf{\grr{}}: Chris Williams (Wide Receiver) $\rightarrow$ Miami Dolphins \ding{55}
      \\\hline 
      \caption{Case study of \name{} and \grr{} retrieval. Term-based retrieval approaches (\abr{tf-idf} used by \grr{})
             is unable to distinguish two players with same name. 
             \name{} correctly 
             identifies the question entity.\label{tb:case}\vspace{-0.3cm}
             }
  \label{tb:case}

  \end{tabular}
\end{table}


\section{Related Work}
\label{sec:related}

Extracting multiple pieces of evidence automatically has applications
from solving crossword puzzles~\cite{littman2002probabilistic}, graph
database construction~\cite{de2009towards}, and understanding
relationships~\cite{chang-09c,iyyer-16} to question
answering~\cite{ferrucci2010building}, which is the focus of this
work.

Given a complex question, researchers have investigated multi-step
retrieval techniques to find an evidence chain. Knowledge graph
question answering approaches~\cite[\emph{inter
    alia}]{talmor-berant-2018-web, zhang2018variational} directly
search the evidence chain from the knowledge graph, but falter when
\abr{kg} coverage is sparse. With the release of large-scale
datasets~\cite{yang+18b}, recent systems~\cite[\emph{inter
    alia}]{nie2019revealing, zhaotransxh2020, asai2020learning,
  Dhingra2020Differentiable} use Wikipedia abstracts (the first paragraph of a
Wikipedia page) as the corpus to retrieve the evidence chain.
\citet{Dhingra2020Differentiable} treat Wikipedia as a knowledge graph, where each entity 
is identified by its textual span mentions, while other approaches~\cite{nie2019revealing, zhaotransxh2020}
directly retrieve passages. They first adopt a single-step retrieval to select the first hop passages 
(or entity mentions), then find the 
next hop candidates directly from Wikipedia links and rerank them.
%
%
%
Like \name{}, \citet{asai2020learning} use beam search to find the
chains but still rely on a graph neural network over Wikipedia links.
\name{} retrieves evidence chains through dense
representations without relying on the corpus semi-structure.
\citet{qi2019answering, qi2020retrieve} iteratively generate the query from the 
question and retrieved history, and use traditional sparse \abr{ir}
systems to select the passage, which complements \name{}'s approach.

\section{Conclusion}
\label{sec:conclu}

We introduce a simple yet effective multi-step dense retrieval method,
\name{}.
By conducting
beam search and globally refreshing negative chains during
training,  \name{} finds reasoning chains in dense
space.
\name{} is competitive to more complex \abr{sota}  
systems albeit not using semi-structured information.
%


While \name{} can uncover relationship embedded within a single
question, future work should investigate how to use these connections
to resolve ambiguity in the question~\cite{elgohary-19,min-20},
resolve entity mentions~\cite{guha-etal-2015-removing}, connect concepts across
modalities~\cite{lei-18}, or to connect related questions to each
other~\cite{elgohary-18}.

\section*{Acknowledgments}

We thank the anonymous reviewers
and meta-reviewer
for their suggestions and comments.
Zhao is supported by the Office of the Director of National Intelligence (\abr{odni}), 
Intelligence Advanced Research Projects Activity (\abr{iarpa}), via the \abr{better} Program contract 2019-19051600005. 
Boyd-Graber is supported by \abr{nsf} Grant IIS-1822494.
Any opinions, findings, conclusions, or recommendations
expressed here are those of the authors and do not
necessarily reflect the view of the sponsors.

\clearpage


\bibliographystyle{style/acl_natbib_2019}
\bibliography{bib/journal-full,bib/multi_ance}

\end{document}